%% file: main.tex
\PassOptionsToClass{a4paper}{IEEEtran}
\def\PaperForArxiv{1}
\RequirePackage[T1]{fontenc}
\documentclass[conference]{IEEEtran}
\usepackage{amsmath,amssymb,bm}
\usepackage{booktabs,array,graphicx}
\usepackage{tikz}
\usetikzlibrary{arrows.meta,positioning}
\usepackage{cite}
\usepackage{float}
\usepackage[hidelinks]{hyperref}
\hypersetup{pdftitle={Consistency Learning and Distillation for One-Step MIMO-OFDM Channel Estimation},pdflang={en}}
\newcommand{\E}{\mathbb{E}}
\newcommand{\sg}{\operatorname{sg}}
\newcommand{\CN}{\mathcal{CN}}
\newcommand{\norm}[1]{\left\lVert#1\right\rVert}

\title{Consistency Learning and Distillation for One-Step MIMO-OFDM Channel Estimation}
\ifdefined\PaperAnonymousReview
\hypersetup{pdfauthor={}}
\author{\IEEEauthorblockN{Anonymous authors}}
\else
\hypersetup{pdfauthor={Fredrik Nguyen and Qiwen Wang}}
\author{\IEEEauthorblockN{Fredrik Nguyen}
\IEEEauthorblockA{KTH Royal Institute of Technology\\Stockholm, Sweden\\\href{mailto:fnguyen@kth.se}{fnguyen@kth.se}}
\and
\IEEEauthorblockN{Qiwen Wang}
\IEEEauthorblockA{Stockholm, Sweden\\\href{mailto:wangqiwen2008@gmail.com}{wangqiwen2008@gmail.com}}}
\fi
\ifdefined\PaperForArxiv
\hypersetup{pdfsubject={Author-prepared preprint}}
\else
\hypersetup{pdfsubject={Conference manuscript}}
\fi

\begin{document}
\maketitle
\raggedbottom
\ifdefined\PaperForArxiv
\pagestyle{plain}
\thispagestyle{plain}
\fi

\begin{abstract}
Iterative neural channel estimators can improve recovery under pilot contamination, but repeated network evaluations increase inference cost. This paper develops observation-anchored consistency and Consistency Mid-Training (CMT) estimators for one-step channel estimation in multiple-input multiple-output orthogonal frequency-division multiplexing (MIMO-OFDM) systems. Both estimators refine pilot-derived least-squares channel estimates in the delay-beam domain. Consistency learning couples endpoint predictions along noisy-to-clean paths. CMT then refines a compact consistency-trained student to predict a flow-matching teacher's endpoint from intermediate trajectory states. Across the evaluated SNR range, the compact CMT student reduces uncoded symbol error rate (SER) by 9.6--20.2\% relative to the one-step flow-matching baseline (0.44--0.98 dB in logarithmic SER reduction), while using 74.5\% fewer parameters and 73.0\% fewer network multiply--accumulates.
\end{abstract}


\section{Introduction}
Accurate channel state information (CSI) is required for coherent detection in MIMO-OFDM systems. Pilot-aided estimation is impaired by thermal noise and by interference from users transmitting overlapping pilots. Least-squares (LS) estimation provides a simple observation~\cite{liu2014survey}, but does not separate the desired channel from pilot contamination. Statistical estimators and learned channel priors can improve recovery by exploiting structure across frequency and space~\cite{liu2014survey,arvinte2023score}.

Generative approaches provide flexible channel priors, yet their inference procedures often require repeated neural evaluations. Flow matching learns a time-dependent velocity field~\cite{lipman2023fm}, and has been applied to channel restoration and plug-and-play estimation~\cite{liu2025fm,zhang2026fm}. One-step channel estimation is already possible with average-velocity learning~\cite{jiang2026avf}, while consistency priors have been incorporated into iterative channel recovery~\cite{li2026cm}. We compare endpoint consistency and teacher-trajectory supervision with one-step flow matching for detection, using matched-backbone and regression controls.

Building on consistency learning~\cite{song2023cm,song2024ict} and Consistency Mid-Training (CMT)~\cite{hu2026cmt}, we develop channel estimators that refine pilot-contaminated LS estimates with one neural function evaluation (NFE). Our contributions are:

\begin{itemize}
\item We develop an observation-anchored endpoint-consistency estimator for pilot-contaminated MIMO-OFDM channels, enforcing agreement between endpoint predictions at adjacent states along observation-to-clean paths.
\item A compact teacher-trajectory student, compared with continued consistency training from shared initial weights and an equal student epoch budget.
\item Matched-backbone comparisons with standard FM and a regression control, reporting normalized mean squared error (NMSE), data-symbol error rate (SER), and estimator parameter and multiply--accumulate (MAC) counts.
\end{itemize}

\section{Related Work}
Recent generative approaches learn channel distributions to support estimation from noisy or contaminated pilots. Arvinte and Tamir use score-based priors for iterative MIMO channel recovery~\cite{arvinte2023score}. Fesl et al.\ use a compact angular-domain network with SNR-matched deterministic reverse diffusion~\cite{fesl2024diffusion}. Zhou et al.\ combine a diffusion prior and approximate likelihood scores for pilot-limited recovery~\cite{zhou2025diffusion}.

More recently, Liu et al.\ learn a noisy-to-clean channel flow~\cite{liu2025fm}, while Zhang and Nadeem use flow-derived denoising in plug-and-play proximal gradient descent~\cite{zhang2026fm}. Zhao et al.\ combine range--null space decomposition with noise-aware iterative flow refinement under latency constraints~\cite{zhao2026null}. Jiang et al.\ learn an average velocity field for one-step estimation~\cite{jiang2026avf}. Diff-Rx embeds one-step conditional diffusion in an iterative receiver, using soft data-symbol feedback to refine subsequent channel estimates~\cite{zhou2026diffrx}.

Another line of research uses consistency models to map trajectory states to a common endpoint~\cite{song2023cm,song2024ict}. Li et al.\ use a consistency prior in an iterative MIMO estimator that alternates pilot-observation updates with angular-domain denoising~\cite{li2026cm}. CMT trains a student to predict a teacher-generated endpoint from intermediate states along the teacher's solver trajectory~\cite{hu2026cmt}.

\section{System Model and Problem Formulation}
Bold lowercase and uppercase letters denote vectors and matrices, respectively. The norms $\norm{\cdot}_2$ and $\norm{\cdot}_F$ denote the Euclidean and Frobenius norms.

\subsection{Received Signal and Pilot Projection}
Consider a multiuser MIMO-OFDM uplink with one desired user transmitting a single spatial stream, four single-stream interferers, $N_r=64$ receive ports and $K=192$ subcarriers. At subcarrier $k$ and OFDM symbol $\ell$, the received vector is
\begin{equation}
 \bm y_{k,\ell}=\bm H_{k,\ell}x_{k,\ell}
 +\sum_{i=1}^{4}\bm H^{(i)}_{k,\ell}x^{(i)}_{k,\ell}
 +\bm n_{k,\ell},
 \label{eq:rx}
\end{equation}
where $x_{k,\ell}$ is the transmitted symbol and $\bm n_{k,\ell}\sim\CN(\bm 0,\bm I)$ denotes normalized Gaussian noise. The desired-user channel $\bm H_{k,\ell}\in\mathbb C^{N_r\times1}$ includes power scaling. With normalized underlying channels $\overline{\bm H}$, define $\bm H=\sqrt{\gamma}\,\overline{\bm H}$ and $\bm H^{(i)}=\sqrt{\eta_i}\,\overline{\bm H}^{(i)}$. Here, $\gamma$ and $\eta_i$ control the desired-signal and interference powers relative to the normalized noise power.

Each slot contains 14 OFDM symbols: two carry demodulation reference signals (DMRS) and twelve carry data. DMRS occupy every other subcarrier (Comb-2), with 96 occupied subcarriers per pilot symbol. Pilot power is normalized to unit average over the complete frequency grid, including unoccupied comb positions. On an occupied resource element, dividing the received vector by the known desired pilot $x_{k,\ell}=p_{k,\ell}\ne0$ gives the LS estimate
\begin{equation}
 \widehat{\bm H}^{\rm LS}_{k,\ell}
 =\bm H_{k,\ell}+\sum_{i=1}^{4}\rho^{(i)}_{k,\ell}\bm H^{(i)}_{k,\ell}
 +\widetilde{\bm n}_{k,\ell},\label{eq:ls}
\end{equation}
where $\rho^{(i)}_{k,\ell}=p^{(i)}_{k,\ell}/p_{k,\ell}$ and $\widetilde{\bm n}_{k,\ell}=\bm n_{k,\ell}/p_{k,\ell}$. Interferers on the other comb do not overlap the desired pilot subcarriers, giving $\rho^{(i)}_{k,\ell}=0$. Two interferers contribute to each LS estimate in this configuration. Pilot-induced noise scaling is absorbed into the effective SNR. The clean target is the desired channel $\bm H$.

\subsection{Representation and Receiver Interface}
Multipath energy often concentrates over a limited region of delays and spatial frequencies. You et al.\ relate angle-delay channel power to wideband array-channel statistics~\cite{you2016acquisition}. Fourier angle-delay representations are also used in learned CSI recovery~\cite{wen2018csinet}. We use a fixed transform to expose this structure to the estimator while retaining all delay-beam coefficients.

We apply a 128-point orthonormal inverse DFT across frequency and a two-dimensional spatial DFT within each polarization's $4\times8$ array. Zero-padding the 96 occupied subcarriers and stacking the real and imaginary parts of both DMRS-symbol estimates gives a real $4\times128\times64$ input tensor: two DMRS symbols times two real components. The clean training target has the same shape. We denote the transformed LS estimate and clean target by $\bm h_0$ and $\bm h_1$, using vector notation in the learning objectives. The network operates on the tensors and also receives context $c=\log(\operatorname{std}(\bm h_0)+10^{-6})$, computed over all observation entries.

\begin{figure}[t]
\centering
\begin{tikzpicture}[node distance=3mm and 3mm,
 box/.style={draw,rounded corners=1pt,align=center,font=\scriptsize,minimum height=9mm,text width=22mm},
 arr/.style={-{Stealth[length=1.3mm]},thin}]
\node[box] (pilot) {Received pilots\\LS projection};
\node[box,right=of pilot] (db) {Delay-beam\\representation};
\node[box,right=of db] (model) {Endpoint model\\one NFE};
\node[box,below=of model] (inv) {Inverse transform\\interpolation};
\node[box,left=of inv] (detect) {Fixed LMMSE\\detector};
\node[box,left=of detect] (ser) {QPSK decisions\\symbol error rate};
\draw[arr] (pilot)--(db); \draw[arr] (db)--(model);
\draw[arr] (model)--(inv); \draw[arr] (inv)--(detect); \draw[arr] (detect)--(ser);
\end{tikzpicture}
\caption{One-step estimation and evaluation chain. The network processes both pilot symbols in one call. All estimators use the same oracle interference-plus-noise covariance in the detector.}
\label{fig:pipeline}
\end{figure}
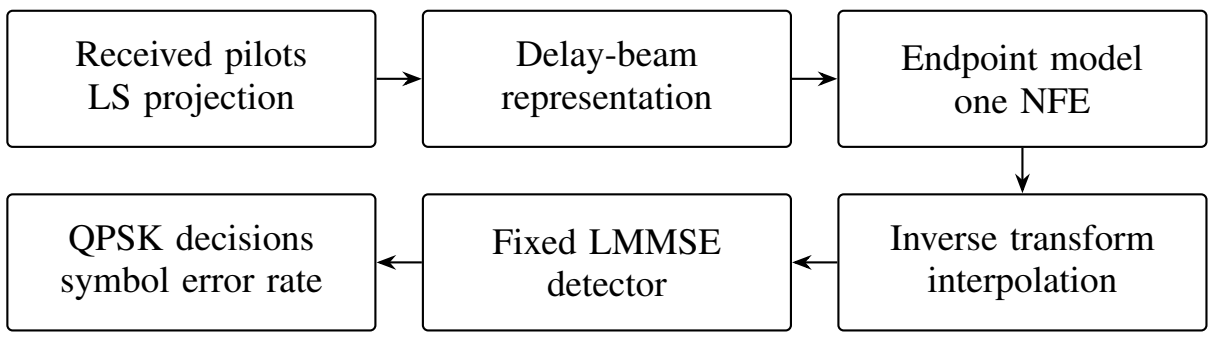

After inversion, linear frequency interpolation restores the full channel grid, with edge copying at the band limits. For each subcarrier and receive port, the two pilot-symbol channel estimates are averaged to obtain one estimate used for all twelve data symbols. All estimates feed the same linear minimum mean square error (LMMSE) detector. Only the channel-estimation module changes across methods (Fig.~\ref{fig:pipeline}).

\section{Endpoint Consistency and CMT}
\subsection{Standard Flow-Matching Reference}
The FM baseline follows the noisy-to-clean channel-restoration construction of Liu et al.~\cite{liu2025fm}, grounded in conditional flow matching~\cite{lipman2023fm}. We adapt this construction to pilot-contaminated delay-beam observations and the U-Net backbones studied here.

For a transformed LS estimate and its paired clean channel $(\bm{h}_0,\bm{h}_1)$, use the linear path
\begin{equation}
 \bm{h}_t=(1-t)\bm{h}_0+t \bm{h}_1,\qquad t\in[0,1].\label{eq:path}
\end{equation}
The conditional path velocity is $\bm{u}=\bm{h}_1-\bm{h}_0$. FM and direct regression use the same adaptively weighted squared-error criterion~\cite[Sec.~4.3 and App.~B.2]{geng2025meanflow}. The weighting reduces the relative influence of training examples with large prediction errors. For the per-example FM error $\ell_{\rm FM}=\norm{v_\phi(\bm{h}_t,t,c)-\bm{u}}_2^2/D$, where $D$ is the number of real entries per example, the objective is
\begin{equation}
 \mathcal L_{\rm FM}=\E\left[\frac{\ell_{\rm FM}}
 {(\sg[\ell_{\rm FM}]+\epsilon)^p}\right],
 \label{eq:fm}
\end{equation}
where $\sg$ denotes stop-gradient, $p$ controls the weighting strength, and $\epsilon>0$ stabilizes the denominator. Weighted per-example losses are averaged over the batch. Regression uses the same criterion with per-example error $\ell_{\rm reg}=\norm{g_\psi(\bm h_0,c)-\bm h_1}_2^2/D$. FM training times follow a logit-normal distribution (see the \hyperref[app:training]{Appendix}). A one-step Euler estimate is
\begin{equation}
 \widehat{\bm{h}}_{\rm FM}=\bm{h}_0+v_\phi(\bm{h}_0,0,c).\label{eq:fm1}
\end{equation}
The teacher uses repeated Euler updates on a uniform time grid.

The identity $\bm{h}_t+(1-t)\bm{u}=\bm{h}_1$ connects velocity and endpoint predictions on this path. The consistency objectives below couple endpoint predictions across path states.

\subsection{Observation-Anchored Endpoint Consistency}
We formulate an observation-anchored endpoint-consistency estimator for one-step channel restoration, building on the boundary-preserving parameterization and training principles of consistency models~\cite{song2023cm,song2024ict}:
\begin{equation}
 f_\theta(\bm{h},t,c)=c_{\rm skip}(t)\bm{h}+c_{\rm out}(t)F_\theta(\bm{h},1-t,c).
\end{equation}
For a fixed preconditioning scale $\sigma_d>0$, the coefficients are
\begin{equation}
\begin{aligned}
 c_{\rm skip}(t)&=\frac{\sigma_d^2}{(1-t)^2+\sigma_d^2},\\
 c_{\rm out}(t)&=\frac{(1-t)\sigma_d}{\sqrt{(1-t)^2+\sigma_d^2}}.
\end{aligned}
\end{equation}
These enforce $f_\theta(\bm{h},1,c)=\bm{h}$ at the clean boundary. As in FM, $t=0$ denotes the noisy observation and $t=1$ the clean channel.

For adjacent levels $t_a<t_b$, form both states using the same pair in~\eqref{eq:path}. The endpoint-consistency objective is
\begin{align}
 \mathcal L_{\rm EC}
 &=\E\left[\rho_\delta\left(f_\theta(\bm{h}_{t_a},t_a,c)
 -\sg[f_\theta(\bm{h}_{t_b},t_b,c)]\right)\right],\label{eq:ec}\\
 \rho_\delta(\bm{z})&=\sqrt{\norm{\bm{z}}_2^2+\delta^2}-\delta.
\end{align}
Here, $\delta>0$ is the loss smoothing parameter. The batch-averaged loss and detached current-weight target follow improved consistency training (iCT)~\cite[Secs.~3.2--3.3]{song2024ict}. Adjacent states are sampled on paired observation-to-clean paths using a progressively refined grid, adapting iCT's exponential interval-growth curriculum~\cite[Sec.~3.4]{song2024ict}. A logit-normal draw selects an adjacent pair, with the higher-$t$ branch supplying the detached target in~\eqref{eq:ec}. The \hyperref[app:training]{Appendix} specifies the sampling schedule. At inference,
\begin{equation}
 \widehat{\bm{h}}_{\rm EC}=f_\theta(\bm{h}_0,0,c),\label{eq:ec1}
\end{equation}
requiring one NFE.

\subsection{CMT with a Compact Endpoint Student}
The student is initialized from a consistency-trained checkpoint. Following CMT~\cite{hu2026cmt}, we freeze a standard-FM teacher and construct its trajectory from the observation:
\begin{equation}
 \bm{z}_0=\bm{h}_0,\qquad \bm{z}_{m+1}=\bm{z}_m+\frac1M v_\phi(\bm{z}_m,m/M,c),
 \label{eq:teacher}
\end{equation}
for $m=0,\ldots,M-1$. We sample a nonterminal state $\widetilde{\bm{z}}_t$ uniformly from a grid along the interpolated teacher trajectory. Following CMT's trajectory-to-endpoint objective~\cite[Sec.~3.2]{hu2026cmt}, we use an MSE discrepancy:
\begin{equation}
 \mathcal L_{\rm CMT}
 =\E_{\bm{h}_0,t}\left[\frac{1}{D}
 \norm{f_\theta(\widetilde{\bm{z}}_t,t,c)-\sg[\bm{z}_M]}_2^2\right].
 \label{eq:cmt}
\end{equation}
The loss is averaged over batch and vector entries. At inference, the student maps the noisy observation directly to an endpoint using~\eqref{eq:ec1}. Intermediate teacher states are used only during training.

\section{Simulation Setup}
\subsection{Channel Data}
Training channels are generated using QuaDRiGa v2.8.1~\cite{jaeckel2014quadriga,quadriga281manual}, based on the 3GPP TR 38.901 urban-macro non-line-of-sight scenario~\cite{3gpp38901}. Table~\ref{tab:data} summarizes the physical configuration. Each UE port is treated as a separate channel example. The corpus contains 816,000 snapshot/port examples, split 99/1 for training and validation. Samples from the same trajectory can occur in both partitions. Channel blocks are power-normalized jointly across antennas and subcarriers, with the first-snapshot scale reused across snapshots.

Receiver evaluation uses a separate CDL-B test set~\cite[Sec.~7.7.1]{3gpp38901} with the physical settings in Table~\ref{tab:data}. It contains 1,000 transmission time intervals (TTIs), with a fixed desired user and four interferers.

\begin{table}[t]
\centering\footnotesize
\caption{Channel and transmission configuration.}\label{tab:data}
\begin{tabular}{@{}>{\raggedright\arraybackslash}p{.46\linewidth}>{\raggedright\arraybackslash}p{.48\linewidth}@{}}
\toprule
Parameter & Value\\\midrule
Carrier frequency & 3.5 GHz\\
Subcarriers; spacing & 192; 30 kHz\\
Sampled bandwidth & 5.76 MHz\\
Base-station array & $4\times8$ dual-polarized panels (64 ports)\\
UE antenna ports & 2, cross-polarized\\
Delay-spread reference & 363 ns\\
UE speed & 1 km/h\\
Desired users; interfering users & 1; 4\\
OFDM symbols per slot & 14\\
Pilot symbols; occupied subcarriers & 2; 96 per symbol (Comb-2)\\
Pilot/data constellation & QPSK\\
\bottomrule
\end{tabular}
\end{table}

\subsection{Training Protocol}
Each training snapshot is held constant across a synthesized slot. Four interference channels are sampled with replacement from the desired example's partition. During training, QPSK pilot realizations are drawn from pilot banks. Receiver evaluation uses fixed pilot realizations. Table~\ref{tab:training} lists training settings and operating points.

Regression, FM, and direct consistency train uninterrupted for ten epochs. CMT starts from the same epoch-five consistency-model weights as the continued-consistency control and receives five CMT epochs, giving both ten total student epochs. The post-training variant adds five consistency epochs using~\eqref{eq:ec} after CMT. These budgets exclude teacher training and trajectory generation.

AdamW uses a constant base learning rate, with a brief initial warmup for regression, FM, and consistency. The weighting exponent $p=1$ was chosen after comparing $p\in\{0,1\}$, where $p=0$ gives unweighted MSE. Other hyperparameters were chosen through manual experimentation. Regression, FM, and consistency use the lowest-validation-NMSE weights among the non-averaged and exponential moving average (EMA) variants evaluated.

FM time sampling and consistency interval selection use the Gaussian parameters in Table~\ref{tab:training}. The Appendix specifies their sampling schedules.

CMT linearly interpolates the frozen teacher's ten-step trajectory onto a uniform grid of 40 intervals. One nonterminal grid state per example is supplied to the student, using a common sampled time within each batch. The teacher endpoint is its target.

Training and validation use pilot-derived channel examples. Receiver evaluation additionally measures data-symbol detection. All estimators use one network evaluation, except for the ten-step teacher reference.

\begin{table}[t]
\centering\footnotesize
\caption{Training settings and operating points.}\label{tab:training}
\begin{tabular}{@{}>{\raggedright\arraybackslash}p{.46\linewidth}>{\raggedright\arraybackslash}p{.48\linewidth}@{}}
\toprule
Parameter & Value\\\midrule
Optimizer; training batch size & AdamW; 64\\
Base learning rate; weight decay & $2\times10^{-4}$; $10^{-4}$\\
Warmup & 1\% of updates\\
Residual weighting $(p,\epsilon)$ & $(1,10^{-4})$\\
Consistency $(\sigma_d,\delta)$ & $(1.6397,10^{-4})$\\
FM time Gaussian $(\mu,s)$ & $(-0.4,1)$\\
Consistency time Gaussian $(\mu,s)$ & $(1.1,2)$\\
\midrule
\multicolumn{2}{@{}l}{\emph{Training observations}}\\
SNR (dB) & Truncated $\mathcal N(-10,2.5^2)$ on $[-16,-4]$\\
INR (dB) & Independent uniform $[0,8]$\\
\midrule
\multicolumn{2}{@{}l}{\emph{Validation and receiver evaluation}}\\
SNR (dB) & $-14,-12,-10,-8,-6$\\
Receiver INR (dB) & $5.229,5.229,5.229,1.0914$\\
\bottomrule
\end{tabular}
\end{table}

\begin{table}[t]
\centering\footnotesize
\caption{Backbones and cost per network evaluation in GMAC ($10^9$ MACs) for four-channel $128\times64$ inputs. Block counts are per resolution level.}\label{tab:backbones}
\setlength{\tabcolsep}{3pt}
\begin{tabular}{@{}lrrr@{}}
\toprule
Setting & U-Net & U-Net-Lite & U-Net-Mini\\\midrule
Base width & 32 & 16 & 8\\
Channel multipliers & $[1,2,4]$ & $[1,1,2]$ & $[1,1,2]$\\
Encoder blocks & 2 & 1 & 1\\
Decoder blocks & 3 & 2 & 2\\
\midrule
Parameters & 3.95 M & 193 k & 49.3 k\\
GMAC & 4.68 & 0.330 & 0.0892\\
\bottomrule
\end{tabular}
\end{table}

\begin{table*}[t]
\centering\footnotesize
\caption{Uncoded SER at all five evaluated SNRs. Bold values mark the lowest one-step SER at the displayed precision, including ties. Calculations use unrounded values. The ten-step teacher is shown separately.}\label{tab:main}
\input{ser_table.tex}
\end{table*}

\subsection{Network Architecture and Reference Estimators}
All backbones adapt Liu et al.'s residual U-Net for channel estimation~\cite{liu2025fm}, with encoder--decoder skip connections~\cite{ronneberger2015unet} and time/context embeddings. All learned estimators receive the input's log-standard-deviation context through the scalar-conditioning embedding. Residual blocks use two $3\times3$ convolutions with GroupNorm and SiLU. All variants include single-head bottleneck attention. The large U-Net also applies attention at feature-map height 32. Compact variants reduce width and depth. Table~\ref{tab:backbones} reports parameter counts and GMAC, including convolutional and linear layers and both attention matrix products.

We evaluate the proposed consistency estimators using two network architectures, U-Net-Lite and U-Net-Mini, and compare their performance against standard FM. As an additional control, direct regression without the FM framework is implemented using U-Net-Lite. For CMT, a large U-Net serves as the teacher, while U-Net-Mini is employed as the student. The teacher model is evaluated with both one and ten Euler steps. One-step FM with U-Net-Lite serves as the baseline reference for evaluating the SER reduction achieved by the compact CMT student.

\begin{figure*}[t]
\centering
\includegraphics[width=.96\textwidth]{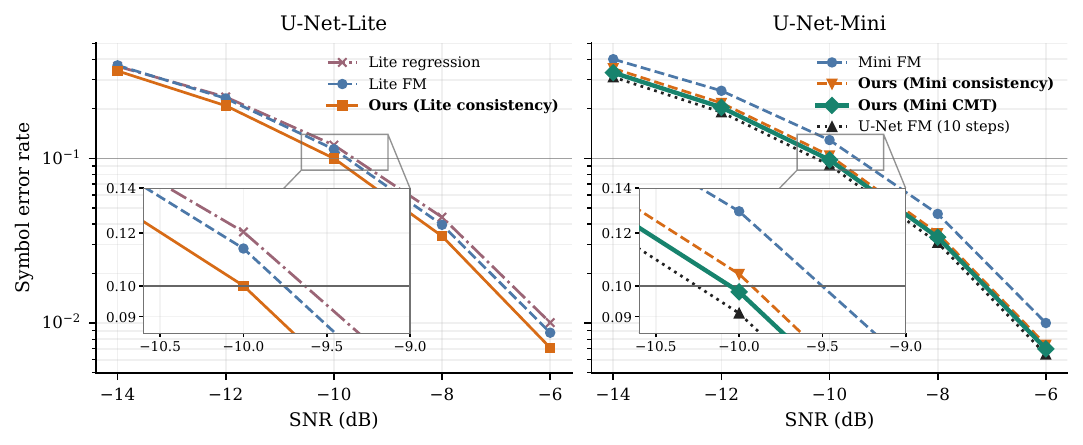}
\caption{Uncoded SER of regression, FM, and our consistency/CMT estimators. Insets magnify the boxed regions using the same logarithmic SER scale. Markers denote measured points.}
\label{fig:ser}
\end{figure*}

\subsection{Receiver and Metrics}
The receiver forms an oracle interference-plus-noise covariance from the true scaled channels. With $\bm H_{\rm intf}=[\bm{H}^{(1)},\ldots,\bm{H}^{(4)}]$, it averages $\bm H_{\rm intf}\bm H_{\rm intf}^{H}+\bm I$ over all OFDM symbols and contiguous windows of 48 subcarriers. For unit-average-power data symbols and estimated desired channel $\widehat{\bm{H}}$, the LMMSE output is
\begin{equation}
 \widehat x=\left(\widehat{\bm{H}}^{H}\bm R_{uu}^{-1}\widehat{\bm{H}}+\bm I\right)^{-1}
 \widehat{\bm{H}}^{H}\bm R_{uu}^{-1}\bm y.\label{eq:detector}
\end{equation}
Nearest-neighbor QPSK decisions give uncoded SER $N_{\rm err}/N_{\rm sym}$ over desired-user data symbols, excluding DMRS. SER is the primary performance metric because it directly measures downstream data-symbol recovery with the common detector. NMSE provides a complementary measure of channel reconstruction accuracy.

For Mini CMT relative to one-step Lite FM at the same evaluated SNR, the SER reductions in percent and dB are
\begin{equation}
\begin{aligned}
 R_{\%}&=100\left(1-\frac{\mathrm{SER}_{\rm CMT}}{\mathrm{SER}_{\rm FM}}\right),\\
 R_{\rm dB}&=10\log_{10}\frac{\mathrm{SER}_{\rm FM}}{\mathrm{SER}_{\rm CMT}}.
\end{aligned}
\label{eq:ser_reduction}
\end{equation}
The latter is a logarithmic SER reduction.

Channel reconstruction is evaluated on all occupied Comb-2 pilot subcarriers. For reference and estimated channel matrices $\bm{H}_n$ and $\widehat{\bm{H}}_n$ in TTI $n$, the pooled NMSE is
\begin{equation}
 \operatorname{NMSE}_{\rm dB}=10\log_{10}
 \frac{\sum_{n=1}^{N_{\rm TTI}}\norm{\widehat{\bm{H}}_n-\bm{H}_n}_F^2}
 {\sum_{n=1}^{N_{\rm TTI}}\norm{\bm{H}_n}_F^2}.
 \label{eq:nmse}
\end{equation}
Here, $N_{\rm TTI}$ is the number of evaluated TTIs.

\section{Results and Discussion}
\subsection{Consistency Relative to Standard Flow Matching}
Table~\ref{tab:main} and Fig.~\ref{fig:ser} compare receiver performance. With U-Net-Lite, our consistency estimator reduces uncoded SER at $-10$ dB from 0.114 to 0.100, a 12.0\% relative reduction. With U-Net-Mini, SER decreases from 0.129 to 0.104, or 19.4\%. Across all five SNRs, consistency reduces SER by 7.5--19.0\% with U-Net-Lite and 12.1--26.4\% with U-Net-Mini versus matched-backbone FM.

With the same backbone and ten-epoch training budget, Lite consistency achieves lower SER than Lite regression at all five evaluated SNRs. At $-10$ dB, SER decreases from 0.120 to 0.100. Regression learns the direct noisy-to-clean mapping, while consistency trains endpoint predictions at adjacent path states to agree. These comparisons suggest that pathwise consistency may help the learned restoration map preserve channel information useful for detection.

\subsection{CMT Performance and Network Cost}
Mini CMT reduces SER relative to Lite FM from 0.367 to 0.332 at $-14$ dB and from 0.00876 to 0.00699 at $-6$ dB. Across the five evaluated points, relative SER reduction is 9.6--20.2\% (0.44--0.98 dB in logarithmic SER reduction). Mini CMT has the lowest one-step SER at $-12$, $-10$, $-8$, and $-6$ dB. At $-14$ dB, one-step U-Net and CMT are nearly tied. With shared initialization and equal student epochs, CMT reduces Mini consistency SER at $-10$ dB by 5.9\% (0.104 to 0.0980).

Mini CMT uses 74.5\% fewer parameters and 73.0\% fewer network MACs than Lite FM (Table~\ref{tab:backbones}).

CMT originally serves as initialization for flow-map post-training~\cite[Sec.~3.1]{hu2026cmt}. After CMT, five additional consistency epochs increase SER at every evaluated SNR, including from 0.0980 to 0.101 at $-10$ dB.

\subsection{Reconstruction Error and Symbol Detection}
Channel-estimation NMSE measures reconstruction fidelity but need not predict downstream detection error~\cite[Sec.~VI-A]{tugnait2010subblock}. Application-oriented MIMO training therefore considers subsequent equalization and precoding~\cite{katselis2013application}.

\begin{figure}[t]
\centering
\includegraphics[width=.97\linewidth]{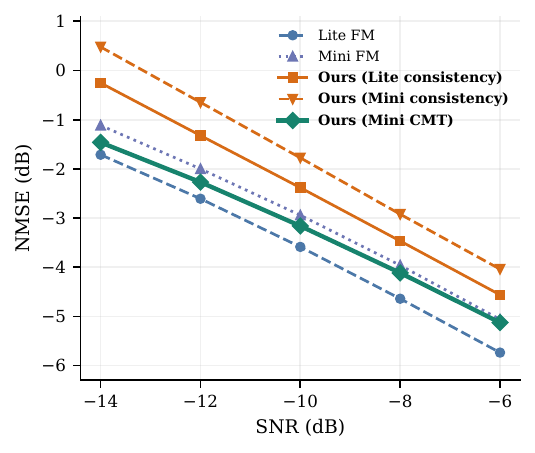}
\caption{NMSE on pilot subcarriers for one-step estimators. Lower values are better.}
\label{fig:nmse}
\end{figure}

In our results (Fig.~\ref{fig:nmse}), direct consistency lowers uncoded SER despite higher NMSE than matched FM at both backbone sizes. At $-10$ dB, Lite FM and Lite consistency have NMSE $-3.59$ and $-2.38$ dB, respectively. Mini CMT reaches $-3.16$ dB, outperforming Mini FM ($-2.94$ dB) but not Lite FM. These rankings hold at all five SNRs.

For illustration, with fixed interference covariance in~\eqref{eq:detector}, positive rescaling of the estimated channel can change NMSE while preserving the detector output's phase and hence QPSK hard decisions.

\section{Conclusion}
Observation-anchored consistency learning and teacher-trajectory distillation enable compact, one-step channel estimation under pilot contamination. With matched backbones, consistency improves downstream uncoded SER over standard flow matching despite higher channel NMSE. Mini CMT reduces uncoded SER by 9.6--20.2\% relative to one-step U-Net-Lite FM while using approximately 75\% fewer parameters and 73\% fewer network MACs. These results concern simulated pilot-contaminated QPSK with oracle interference covariance. Practical covariance estimation and broader propagation conditions remain directions for further evaluation.

\ifdefined\PaperAnonymousReview\else
\newpage
\section*{Acknowledgment and Disclosure}
OpenAI ChatGPT and Codex assisted with drafting and revising text in the Abstract, Sections I–VII, and Appendix, as well as with figure preparation. The authors reviewed and verified the scientific content, methodology, and reported results.
\fi

\appendix[Training Implementation]\label{app:training}
FM samples $t=\operatorname{sigmoid}(Z)$, with $Z\sim\mathcal N(\mu,s^2)$ and the parameters in Table~\ref{tab:training}. For consistency, a uniform grid in $t$ increases from 10 to 1,280 intervals over eight stages defined by optimizer-update count. Let $S$ be the total number of optimizer updates and $j\in\{0,\ldots,S-1\}$ the update index. The number of grid points is
\begin{equation}
\begin{aligned}
 N_j&=1+\min\!\left\{1280,\,10\cdot2^{\lfloor j/K_0\rfloor}\right\},\\
 K_0&=\max\!\left\{1,\left\lfloor S/8\right\rfloor\right\}.
\end{aligned}
\end{equation}
With the consistency Gaussian parameters in Table~\ref{tab:training}, draw $Z\sim\mathcal N(\mu,s^2)$ and select the interval index
\begin{equation}
 i=\min\!\left(N_j-2,\left\lfloor(N_j-1)\operatorname{sigmoid}(Z)\right\rfloor\right).
\end{equation}
Its endpoints are $t_a=i/(N_j-1)$ and $t_b=(i+1)/(N_j-1)$ in~\eqref{eq:ec}. The backbone receives $1-t$.

\bibliographystyle{IEEEtran}
\bibliography{references}
\end{document}

%% file: ser_table.tex
\begin{tabular}{lrrrrr}
\toprule
Estimator & SER at $-14$ dB & SER at $-12$ dB & SER at $-10$ dB & SER at $-8$ dB & SER at $-6$ dB\\\midrule
\multicolumn{6}{l}{\textit{One-step estimators (one network evaluation)}}\\
U-Net FM (1 step) & \textbf{0.332} & 0.206 & 0.100 & 0.0353 & 0.00783 \\
Lite regression & 0.363 & 0.236 & 0.120 & 0.0440 & 0.0101 \\
Lite FM & 0.367 & 0.232 & 0.114 & 0.0396 & 0.00876 \\
Mini FM & 0.401 & 0.258 & 0.129 & 0.0461 & 0.0100 \\
\textbf{Ours (Lite consistency)} & 0.340 & 0.209 & 0.100 & 0.0338 & 0.00709 \\
\textbf{Ours (Mini consistency)} & 0.352 & 0.217 & 0.104 & 0.0352 & 0.00738 \\
\textbf{Ours (Mini CMT)} & \textbf{0.332} & \textbf{0.204} & \textbf{0.0980} & \textbf{0.0333} & \textbf{0.00699} \\
Mini CMT + post-training & 0.343 & 0.211 & 0.101 & 0.0342 & 0.00728 \\
\midrule
\multicolumn{6}{l}{\textit{Iterative teacher reference}}\\
U-Net FM (10 steps) & 0.312 & 0.191 & 0.0910 & 0.0308 & 0.00645 \\
\bottomrule
\end{tabular}